\documentclass{article}

\usepackage{microtype}
\usepackage{graphicx}
\usepackage{booktabs} 
\usepackage{mathtools}
\usepackage{amssymb}
\usepackage{bbm}
\usepackage{todonotes}
\usepackage[T1]{fontenc} 

\usepackage{multirow} 
\usepackage{footnote} 
\graphicspath{ {./images/} }
\usepackage{subcaption}
\usepackage{tikz}
\usetikzlibrary{calc,matrix}
\usepackage{hyperref}

\usepackage[accepted]{icml2018}

\icmltitlerunning{Loss-Calibrated Approximate Inference in Bayesian Neural Networks}

\begin{document}

\twocolumn[
\icmltitle{Loss-Calibrated Approximate Inference in Bayesian Neural Networks}



\icmlsetsymbol{equal}{*}

\begin{icmlauthorlist}
\icmlauthor{Adam D. Cobb}{Eng}
\icmlauthor{Stephen J. Roberts}{Eng}
\icmlauthor{Yarin Gal}{Com}
\end{icmlauthorlist}

\icmlaffiliation{Eng}{Department of Engineering Science, University of Oxford, Oxford, United Kingdom}
\icmlaffiliation{Com}{Department of Computer Science, University of Oxford, Oxford, United Kingdom}

\icmlcorrespondingauthor{Adam D. Cobb}{acobb@robots.ox.ac.uk}

\icmlkeywords{Machine Learning, ICML}

\vskip 0.3in
]



\printAffiliationsAndNotice{} 

\begin{abstract}
Current approaches in approximate inference for Bayesian neural networks minimise the Kullback--Leibler divergence to approximate the true posterior over the weights. However, this approximation is without knowledge of the final application, and therefore cannot guarantee optimal predictions for a given task. 
To make more suitable task-specific approximations, we introduce a new \emph{loss-calibrated} evidence lower bound for Bayesian neural networks in the context of supervised learning, informed by Bayesian decision theory. By introducing a lower bound that depends on a utility function, we ensure that our approximation achieves higher utility than traditional methods for applications that have asymmetric utility functions. 
Furthermore, in using dropout inference, we highlight that our new objective is identical to that of standard dropout neural networks, with an additional utility-dependent penalty term.
We demonstrate our new loss-calibrated model with an illustrative medical example and a restricted model capacity experiment, and highlight failure modes of the comparable weighted cross entropy approach. Lastly, we demonstrate the scalability of our method to real world applications with per-pixel semantic segmentation on an autonomous driving data set.
\end{abstract}

\section{Introduction}

Bayesian neural networks (BNNs) capture uncertainty and provide a powerful tool for making predictions with highly complex input data, in domains such as computer vision \cite{kendall2015bayesian,kendall2017uncertainties} and reinforcement learning \citep{gal2016improving}. Recent applications, which range from diagnosing diabetes \cite{leibig2017leveraging} to using BNNs to perform end-to-end control in autonomous cars \citep{Amini2017}, demonstrate the capabilities available when distributions over predictions are sought. However, these are applications where making a non-optimal prediction might result in a life or death outcome, and incorporating Bayesian decision theory into such applications ought to be a necessity for taking into account asymmetries in how we penalise errors. 

\begin{figure}
    \centering
        \includegraphics[width=\columnwidth]{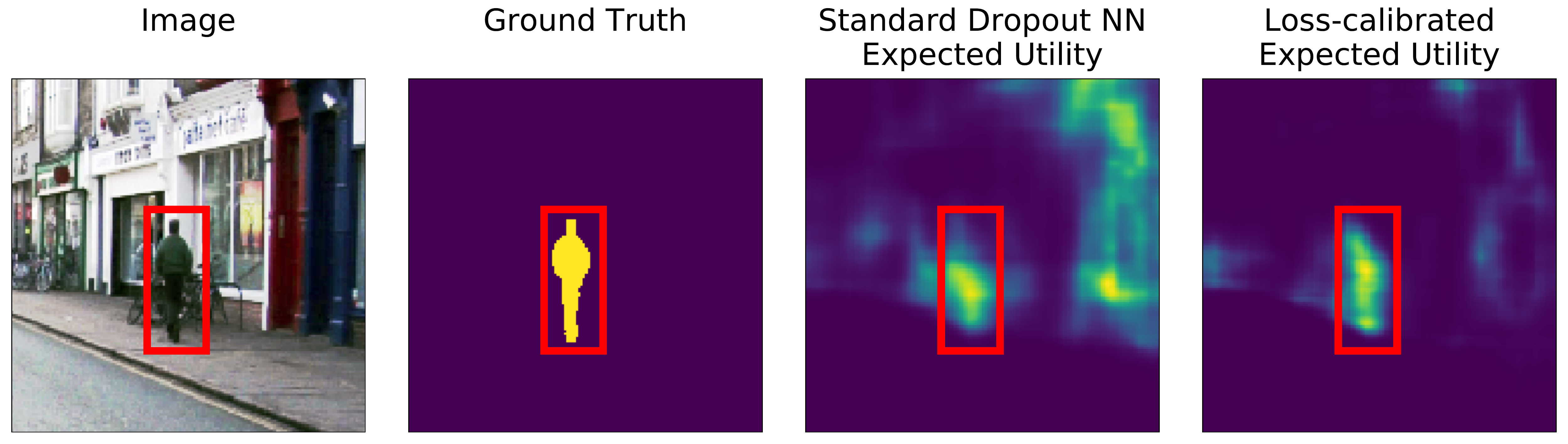}
    \caption{
Per-pixel semantic segmentation (SegNet-Basic \cite{badrinarayanan2017segnet}) trained on an autonomous driving dataset (Camvid \cite{brostow2009semantic}), and its loss-calibrated variant. Our loss-calibrated version achieves higher accuracy on pedestrian and car classes (for which we define the utility to be high) without deteriorating accuracy on background classes. The left-most panel depicts the input image, the next panel depicts the ground truth segmentation for the pedestrian class, and the third and fourth panels depict the expected utility (calculated using Equation \eqref{eq:exp_error} below) for the standard SegNet and the loss calibrated variant respectively (brighter inside the red box, signifying the pedestrian, is better).    
    }\label{fig:seg_example}
\end{figure}

In many applications, the cost of making an incorrect prediction may vary depending on the nature of the mistake. For example, in the diagnosis of a disease, doctors must carefully combine their confidence in their diagnosis with the cost of making a mistake.
Such tasks have a clear asymmetric cost in making false predictions. The cost of falsely diagnosing a disease when a patient does not have it (\emph{false positive}) may be orders of magnitudes lower than not diagnosing a disease when it is present (\emph{false negative}). 
The cost of making different kinds of errors, such as comparing false positives to false negatives, is captured by a \emph{utility function} which guides predictions in the presence of uncertainty\footnote{Note that to avoid confusion with deep learning terminology, we avoid from referring to this as a \textit{cost} function, and instead use the term \textit{utility} function.}. As an example, Figure \ref{fig:seg_example} shows how we can use a utility function to encode our preference for making optimal predictions in labelling pedestrians for autonomous driving tasks.
The field of Bayesian decision theory is concerned with making such optimal predictions given specified utility functions \cite{berger1985statistical}. 

Although Bayesian decision theory is not often considered when applying Bayesian methods, it is a framework that seamlessly combines uncertainty with task-specific utility functions to make rational predictions. We can encode any asymmetries into our utility function and rely on the framework of Bayesian decision theory to make a prediction that aims to maximise this utility. Those familiar with Bayesian models often find themselves performing inference through optimising a marginal likelihood term to determine a set of model parameters. A likelihood and prior must be defined and the end goal is the predictive distribution. The process of performing inference to obtain the predictive distribution is almost always task-agnostic. In contrast, in Bayesian decision theory we use a utility function to choose an optimal label in the presence of ambiguous predictions, rather than relying solely on the predictive distribution. We will review Bayesian decision theory in more detail in Section \ref{sec:BDT}.

BNNs require approximate inference, and using Bayesian decision theory with approximate inference is non-trivial \cite{lacoste2011approximate}.
Asymmetric utilities may result in suboptimal predictions when using an approximate inference method that is task-agnostic. 
The act of simply minimising a distance metric between an approximate and a true distribution may achieve a high overall predictive accuracy, but it might not be accurate enough over important regions of the output space (such as regions where we might have high cost for incorrect predictions).
As an example, if an engine has a temperature sensor that we use to predict the possibility of a catastrophic failure, we are predominantly concerned about the accuracy of our model in the space near the temperature threshold. This example led \citet{lacoste2011approximate} to argue that models must be aware of the utility during inference, if they are required to make approximations. 

Rather than following the framework of Bayesian decision theory, modern neural networks rely on hand-crafted losses to enable suitable network weights to be learned. 
As an example, both \citet{mostajabi2015feedforward} and \citet{xu2014tell} adapt the cross entropy loss to overcome class imbalances in segmentation data. They rely on scaling the loss using training data statistics, such as multiplying by the inverse frequency of each class. However, when faced with noisy labels, this approach can lead to severe over-fitting, as we explain in Sections \ref{sec:illust} and \ref{sec:mnist}.
This last failure constitutes a good example for the importance of clearly separating our loss from our utility function: the loss corresponds to a log likelihood, which describes our noise model. Scaling parts of the loss corresponds to placing explicit assumptions over the noise in the data. A utility function, on the other hand, determines the consequences of making an incorrect prediction \citep[Page~21]{rasmussen2006gaussian}. 

In this paper, we extend the framework introduced by \citet{lacoste2011approximate} with recent work on Bayesian neural networks \cite{gal2016dropout} to provide a theoretically sound way of making optimal predictions for real-world learning tasks. We introduce a new evidence lower bound loss for Bayesian neural network inference that incorporates a task-specific utility function, which can be implemented as a novel penalty term added to the standard dropout neural network:
\begin{equation}
\begin{split}
\mathcal{L}
(
&\boldsymbol{\omega},
\mathbf{H}) \propto \underbrace{-\sum_i \log p(\mathbf{y}_i \mid \mathbf{x}_i, \hat{\boldsymbol{\omega}}_i) + \mid\mid\boldsymbol{\omega}\mid\mid^2}_{\text{Equivalent to standard dropout loss}}\\
& \underbrace{ -\sum_i\left( \log \sum_{\mathbf{c} \in C} u\left(\mathbf{h}_i, \mathbf{c}\right) p(\mathbf{y}_i=\mathbf{c} \mid \mathbf{x}_i ,\hat{\boldsymbol{\omega}}_i)\right)}_{\text{Our additional utility-dependent penalty term}}
\end{split}
\end{equation}
with $\hat{\boldsymbol{\omega}}_i$ dropped-out weights, $\mathbf{h}_i$ as the optimal prediction, $\mathbf{c}$ ranging over possible class labels and $u$ as the utility function, where these definitions are provided in Section \ref{sec:LCB}.
We introduce the \emph{loss-calibrated Bayesian neural network} (LCBNN) as a framework for applying utility-dependent approximate variational inference in Bayesian neural networks to result in models that maximise the utility for given tasks. By specifying a utility we gain the additional advantage of making our assumptions about a task interpretable to the user.

Our paper is organised as follows: in Section \ref{sec:theory}, we start by reviewing recent literature on Bayesian neural networks followed by a summary of Bayesian decision theory. In Section \ref{sect:LC} we derive the loss-calibrated evidence lower bound for our model. Section \ref{sec:illust} explains the motivation for our loss through an illustrative example, where accuracy is shown to be an unsuitable performance metric. Our experiment in Section \ref{sec:mnist} shows how the utility function can be used with limited capacity models, where obtaining good performance across all classes is impossible because of the restricted model size, and the utility is used to prioritise certain classes. Our final experiment in Section \ref{sec:segnet} demonstrates our model's ability to scale to larger deeper networks with an autonomous driving application. We offer insights into future work in Section \ref{sec:conc}.

\section{Theory}\label{sec:theory}

In this Section we introduce BNNs and Bayesian decision theory. We then combine them in Section \ref{sec:LCB} to introduce our loss-calibrated Bayesian neural network.

\subsection{Bayesian Neural Networks}\label{sec:BNN}

Bayesian neural networks offer a probabilistic alternative to neural networks by specifying prior distributions over the weights \cite{mackay1992practical, neal1995bayesian}. The placement of a prior $p(\omega_i)$ over each weight $\omega_i$ leads to a distribution over a parametric set of functions. The motivation for working with BNNs comes from the availability of uncertainty in its function approximation, $\mathbf{f}^{\boldsymbol{\omega}}( \mathbf{x})$. In training, we want to infer the posterior over the weights:
\begin{equation}\label{eq:Bayes}
p(\boldsymbol{\omega}\mid  \mathbf{X},\mathbf{Y}) = \frac{p(\mathbf{Y}\mid \boldsymbol{\omega}, \mathbf{X}) p(\boldsymbol{\omega})}{p(\mathbf{Y}\mid \mathbf{X})}. 
\end{equation}
We define the prior $p(\boldsymbol{\omega})$ for each layer $l \in L$ as a product of multivariate normal distributions $\prod_{l=1}^L\mathcal{N}(\mathbf{0},\mathbf{I}/\lambda_l)$ (where $\lambda_l$ is the prior length-scale) and the likelihood $p(\mathbf{y}\mid \boldsymbol{\omega}, \mathbf{x})$ as a softmax for multi-class classification:
\begin{equation}\label{eq:softmax}
p(\mathbf{y} = c_i \mid \boldsymbol{\omega}, \mathbf{x}) = \frac{\exp\{\mathbf{f}_{c_i}^{\boldsymbol{\omega}}( \mathbf{x})\}}{\sum_{c_j}\exp\{\mathbf{f}_{c_j}^{\boldsymbol{\omega}}( \mathbf{x})\}}.
\end{equation}
In testing, the posterior is then required for calculating the predictive distribution $p(\mathbf{y}^*\mid \mathbf{x}^* , \mathbf{X}, \mathbf{Y})$ for a given test point $\mathbf{x}^*$.

Irrespective of whether we can analytically derive the product of the likelihood and the prior, or sample values from this product $p(\mathbf{Y}\mid \boldsymbol{\omega}, \mathbf{X}) p(\boldsymbol{\omega})$, we still need to specify the normalising factor $p(\mathbf{Y}\mid \mathbf{X})$, known as the marginal likelihood. This requires the integration
\begin{equation}\label{eq:ml}
p(\mathbf{Y}\mid \mathbf{X}) = \int_{\boldsymbol{\omega}} p(\mathbf{Y}\mid \boldsymbol{\omega}, \mathbf{X}) p(\boldsymbol{\omega})\mathit{d}\boldsymbol{\omega},
\end{equation}
which is the bottleneck in performing inference in BNNs. At test time, techniques involving variational inference (VI) \citep{jordan1998introduction} replace the posterior over the weights with a variational distribution $q_{\boldsymbol{\theta}} (\boldsymbol{\omega})$, where we have defined our distribution to depend on the variational parameters $\boldsymbol{\theta}$. For our BNN the parameters $\boldsymbol{\omega}$ are weight matrices $\boldsymbol{\theta}$ multiplied by diagonal matrices with Bernoulli distributed random variables on the diagonal (using the dropout approximating distribution)\footnote{For brevity we drop subscript $\boldsymbol{\theta}$ when it is clear from context.}.
Dropping weights during test time is known as Monte Carlo dropout \cite{gal2016uncertainty} and acts as a test-time approximation for calculating the predictive distribution. 

Rather than using the predictive distribution as our starting point, in VI we aim to minimise the KL divergence between our approximate posterior over the network weights and the true posterior \citep[Eq.~2.3]{gal2016uncertainty}. The minimisation of this KL divergence is then reformulated as the maximisation of the evidence lower bound (ELBO), $\mathcal{L}_{\text{VI}}(\theta)$.

\subsection{Bayesian Decision Theory}\label{sec:BDT}
We rely on the framework of Bayesian decision theory for our model\footnote{We are interested in Bayesian decision theory in the context of supervised learning, so we replace terminology referring to `decisions' or `actions' with \textit{optimal predictions}.} (See \citet[Chapter~1 \& Chapter~5]{berger1985statistical} for more details.). The motivation for selecting this framework is due to the way in which it deals with uncertainty. Any prediction we make should involve the uncertainty in our knowledge over the \emph{state of nature}, $\boldsymbol{\omega}$. In our scenario, we can think of our knowledge of the world state as our confidence in the model parameters (i.e. the function explaining the data), which is given by the posterior over the weights in Equation~\eqref{eq:Bayes}. 

In Bayesian decision theory we also introduce the concept of a task-specific utility function, which is a vital part of making optimal predictions. 
Any agent expected to make a prediction for a specific task must be informed as to how their choices are valued. In a binary labelling task, it is intuitive to imagine that an agent may be more concerned about avoiding false negatives than achieving a high accuracy. Therefore a clear way of defining the goal of a task is to define a utility function that captures the way in which predictions are valued. The agent then aims to select the prediction that maximises the expected utility with respect to the posterior over the parameters.

In the Bayesian decision theory literature, the utility is often introduced as a `loss function', where the aim is to minimise the expected loss. We purposely avoid using `loss', so as to clearly distinguish it from the loss referred to in the deep learning literature. Furthermore, in work related to Bayesian decision theory, the expected utility is sometimes used interchangeably with conditional-gain. The parallel can also be seen with the relationship between the expected loss and conditional-risk. We further highlight the similarities between the action-reward paradigm of reinforcement learning \citep[Chapter~1]{sutton1998reinforcement} and the utility received as a consequence of making a prediction.

Another possible avenue for confusion is in the definition of `decision'. There is a potential to mix terminology, such as `action', `decision', `label' and `optimal model output'. For classification in supervised learning using BNNs, we often denote a probability vector output from the network for a given input $\mathbf{x}_i$ as $\mathbf{y}_i$.
Rather confusingly, $\mathbf{y}_i$ is also used to denote the observed label for input $\mathbf{x}_i$, which takes values $\mathbf{c}$ from the space of all possible classes $\mathcal{C}$ (e.g.\ class labels $0-9$ for MNIST experiments \cite{lecun1998gradient})\footnote{In our notation we use a vector $\mathbf{c}$ to allow for multiple model outputs.}. In standard NNs we also use $\mathbf{y}_i$ to refer to the class with the highest probability in the probability vector output from the model. We avoid this here as the optimal prediction might be different to the \textit{argmax}.
To avoid confusion we denote the probability vector output from the model as $\mathbf{f}^{\boldsymbol{\omega}}(\mathbf{x}_i)$. 
The probability vector model output $\mathbf{f}^{\boldsymbol{\omega}}(\mathbf{x}_i)$ can be seen as a `recommendation', where the actual chosen label or action could be different.
We refer to the chosen label prediction for a given input $\mathbf{x}_i$ as $\mathbf{h}_i$, which can take any label assignment $\mathbf{c} \in \mathcal{C}$.  

In Bayesian decision theory, the overall process is divided into two separate tasks: \emph{probabilistic inference} and \emph{optimal label prediction}. Here, we include details of these two tasks.

\subsubsection{Probabilistic Inference}
When we have access to the true posterior, we can think of probabilistic inference as averaging over the model parameters $\boldsymbol{\omega}$ to infer a predictive distribution $p(\mathbf{y}^*\mid \mathbf{x}^* , \mathbf{X}, \mathbf{Y})$, which can be shown as the integration:
\begin{equation}\label{eq:inference}
p(\mathbf{y}^*\mid \mathbf{x}^* , \mathbf{X}, \mathbf{Y}) =  \int_{\boldsymbol{\omega}} p(\boldsymbol{\omega} \mid \mathbf{X}, \mathbf{Y}) p(\mathbf{y}^* \mid \mathbf{x}^*, \boldsymbol{\omega})\mathit{d}\boldsymbol{\omega}.
\end{equation}

\subsubsection{Optimal Label Prediction}

Having inferred the predictive distribution, in the context of supervised learning for classification, we must make a prediction as to what label to assign for a given input $\mathbf{x}^*$. The label we assign both depends on the specific task and the uncertainty.

Therefore we introduce a utility function $u\left( \mathbf{h}=\mathbf{c}, \mathbf{y}^*=\mathbf{c}'\right)$ (or $u\left( \mathbf{c}, \mathbf{c}'\right)$), which defines what we will gain from predicting different labels $\mathbf{h}$.
We note that in practice, we will use the transformed utility function \citep[Page~60]{berger1985statistical}, whereby we bound the utility\footnote{See explanation in Appendix \ref{ap:bound}} to only take positive values, making an assumption that a lower bound is always possible to find.

\section{Loss-Calibrated Approximate Inference in BNNs}
\label{sect:LC}

To combine our uncertainty in our prediction with the task-specific utility function, we average the utility over the predictions $\mathbf{y}^*$ to give the conditional-gain in assigning a label $\mathbf{h}$ conditioned on a test input $\mathbf{x}^*$:
\begin{equation}\label{eq:exp_error}
\begin{split}
&\mathcal{G}(\mathbf{h} = \mathbf{c} \mid \mathbf{x^*})\\ &= \int_{\mathbf{y^*}} u\left(\mathbf{h} = \mathbf{c}, \mathbf{y}^*= \mathbf{c}'\right) p(\mathbf{y^*}= \mathbf{c}' \mid  \mathbf{x^*}, \mathbf{X}, \mathbf{Y}) \mathit{d} \mathbf{c}'
\end{split}
\end{equation}
The label $\mathbf{h}$ that maximises the conditional-gain is defined as the chosen optimal prediction $\mathbf{h}^*$ for the given input $\mathbf{x}^*$
\begin{equation}\label{eq:opt_task_log}
\begin{split}
\mathbf{h}^*(\mathbf{x^*}) &= \underset{ \mathbf{c} \in \mathcal{C}}{\mathrm{argmax}}\ \mathcal{G}(\mathbf{h}= \mathbf{c} \mid\mathbf{x^*})\\
&= \underset{\mathbf{c} \in \mathcal{C}}{\mathrm{argmax}}\ \log \left(\mathcal{G}(\mathbf{h}= \mathbf{c} \mid\mathbf{x^*})\right)
\end{split}
\end{equation}
conditioned on the dataset $\{\mathbf{X,Y}\}$.

We can rewrite our predictive conditional-gain in terms of an integration with respect to $\boldsymbol{\omega}$:
\begin{align}\label{eq:cond_gain}
&\mathcal{G}(\mathbf{h}= \mathbf{c} \mid\mathbf{x^*})
\notag\\
&=\int_{\mathbf{c}'} u\left(\mathbf{c}, \mathbf{c}'\right) p(\mathbf{y^*} = \mathbf{c}' \mid  \mathbf{x^*},\mathbf{X},\mathbf{Y}) \mathit{d} \mathbf{c}'  
\notag\\
&=\int_{\mathbf{c}'} u\left(\mathbf{c}, \mathbf{c}'\right)\int_{\boldsymbol{\omega}} p(\mathbf{y^*} =\mathbf{c}' \mid \boldsymbol{\omega},  \mathbf{x^*}) p(\boldsymbol{\omega} \mid \mathbf{X}, \mathbf{Y}) \mathit{d} \boldsymbol{\omega} \mathit{d}\mathbf{c}'
\notag\\
&=\int_{\boldsymbol{\omega}} \left[\int_{\mathbf{c}'} u\left(\mathbf{c}, \mathbf{c}'\right) p(\mathbf{y^*}=\mathbf{c}' \mid \boldsymbol{\omega},  \mathbf{x^*})  \mathit{d}\mathbf{c}' \right]  p(\boldsymbol{\omega} \mid \mathbf{X}, \mathbf{Y}) \mathit{d} \boldsymbol{\omega}
\notag\\
&=\int_{\boldsymbol{\omega}} \mathcal{G}(\mathbf{h} = \mathbf{c} \mid\mathbf{x^*},\boldsymbol{\omega})
 p(\boldsymbol{\omega} \mid \mathbf{X}, \mathbf{Y}) \mathit{d} \boldsymbol{\omega},
\end{align}
with the definition 
\begin{equation}
\mathcal{G}(\mathbf{h} = \mathbf{c} \mid\mathbf{x^*},\boldsymbol{\omega}) : =\int_{\mathbf{c}'} u\left(\mathbf{c}, \mathbf{c}'\right) p(\mathbf{y^*}=\mathbf{c}' \mid \boldsymbol{\omega},  \mathbf{x^*})  \mathit{d}\mathbf{c}'.\notag
\end{equation}
For example, if the likelihood is the categorical-softmax defined in Equation \eqref{eq:softmax}, with $\mathbf{y^*}$ taking values of possible classes $c$,
then we calculate $\mathcal{G}(\mathbf{h} \mid\mathbf{x^*},\boldsymbol{\omega})$ by averaging the utility $u\left(\mathbf{h}, \mathbf{y^*}=c'\right)$ with respect to all classes $c'\in \mathcal{C}$, weighted by the probability of that class.
On the other hand, if our likelihood were $\mathcal{N}\left(\mathbf{y^*}; \mathbf{f}^{\boldsymbol{\omega}}(\mathbf{x^*}),\boldsymbol{\Sigma}\right)$, as is common for regression, we could use MC sampling to approximate the conditional-gain.

\subsection{Extending the Loss Function of the BNN}\label{sec:LCB}
In introducing Bayesian decision theory, it is now possible to see how approximating the true posterior of a BNN may lead to sub-optimal predictions, in terms of a task-specific utility. We may learn a lower bound that is loose in areas that our utility demands a tighter fit. Therefore we extend the loss function of a BNN by deriving a new lower bound that depends on the network weights and the utility.

We define the marginal conditional-gain $\mathcal{G}(\mathbf{H} \mid\mathbf{X})$ for the entire input data using a conditional independence assumption over our inputs where
\begin{equation}\label{eq:cond_gain_training}
\begin{split}
\mathcal{G}(\mathbf{H} \mid\mathbf{X}) &: = \int_{\boldsymbol{\omega}} \prod_j\mathcal{G}(\mathbf{h}_j \mid\mathbf{x}_j,\boldsymbol{\omega})
 p(\boldsymbol{\omega} \mid \mathbf{X}, \mathbf{Y}) \mathit{d} \boldsymbol{\omega}\\
&: = \int_{\boldsymbol{\omega}} \mathcal{G}(\mathbf{H} \mid\mathbf{X},\boldsymbol{\omega})
 p(\boldsymbol{\omega} \mid \mathbf{X}, \mathbf{Y}) \mathit{d} \boldsymbol{\omega}.\\
 \end{split}
\end{equation}
and where we assume that given the model parameters, the optimal prediction depends only on the input $\mathbf{x}_j$. If this conditional-gain is large, we have assigned high predictive probability to class labels that give a high task-specific utility across our data. Whereas, low values of the conditional-gain imply that our choice of $\mathbf{H}$ has led to an undesirable low task-specific utility over our data. Therefore, given our aim of assigning class labels in a way that maximises the utility, we choose to maximise the conditional-gain. Furthermore, we will show that this is equivalent to minimising a KL divergence between the approximating distribution $q(\boldsymbol{\omega})$ and a \textit{calibrated} posterior, which results in a loss function that is comparable to the BNN loss introduced in Section \ref{sec:BNN}.

In order to maximise the conditional-gain, we must integrate with respect to the parameters $\boldsymbol{\omega}$ and optimise with respect to the optimal predictions $\mathbf{H}$.
However, due to the intractability of the integration, we must define a lower bound to the log conditional-gain, which we maximise instead:
\begin{equation}
\log \left(\mathcal{G}(\mathbf{H} \mid\mathbf{X})\right) \geq \mathcal{L}(q(\boldsymbol{\omega}),\mathbf{H}),
\end{equation}
where we follow the derivation of \citet{lacoste2011approximate}
by introducing the approximate posterior $q(\boldsymbol{\omega})$ and applying Jensen's inequality:
\begin{align}\label{eq:Jensen}
&\log \left(\mathcal{G}(\mathbf{H} \mid\mathbf{X})\right)
\notag\\
&=\log \left(\int_{\boldsymbol{\omega} }q(\boldsymbol{\omega})\frac{
p(\boldsymbol{\omega} \mid \mathbf{X}, \mathbf{Y}) \mathcal{G}(\mathbf{H} \mid\mathbf{X},\boldsymbol{\omega})}{q(\boldsymbol{\omega})}\mathit{d}\boldsymbol{\omega} \right) 
\notag\\
&\geq \int_{\boldsymbol{\omega} }q(\boldsymbol{\omega})\log \left(\frac{
p(\boldsymbol{\omega} \mid \mathbf{X}, \mathbf{Y}) \mathcal{G}(\mathbf{H} \mid\mathbf{X},\boldsymbol{\omega})}{q(\boldsymbol{\omega})}\right)\mathit{d}\boldsymbol{\omega}
\notag\\
&:= \mathcal{L}(q(\boldsymbol{\omega}),\mathbf{H}).
\end{align}
We will show that this lower bound can be approximated well and can be reformulated as the \emph{standard optimisation objective} loss for a BNN with an additional penalty term. However, to gain further insight, we can also view the maximisation of this lower bound as equivalent to the minimisation of the KL divergence (see proof in Appendix \ref{ap:proof}):
\begin{equation}\label{eq:KL}
KL(q\mid\mid \tilde{p}_h) = \log \left(\mathcal{G}(\mathbf{H} \mid\mathbf{X})\right) - \mathcal{L}(q,\mathbf{H}),
\end{equation}
where the probability distribution
\begin{equation}
\tilde{p}_h = \frac{p(\boldsymbol{\omega} \mid \mathbf{X,Y}) \mathcal{G}(\mathbf{H} \mid\mathbf{X},\boldsymbol{\omega})}{\mathcal{G}(\mathbf{H} \mid\mathbf{X})}
\end{equation}
is the true posterior scaled by the conditional-gain. Therefore we are calibrating the approximate posterior to take into account the utility.

We now derive the loss-calibrated ELBO for the BNN by expanding our lower bound:
\begin{equation}\label{eq:expand_LB}
\begin{split}
&\mathcal{L}(q(\boldsymbol{\omega}),\mathbf{H}) \\
&= \underbrace{\int_{\boldsymbol{\omega}} q (\boldsymbol{\omega}) \log p(\mathbf{Y} \mid \mathbf{X}, \boldsymbol{\omega}) \mathit{d}\boldsymbol{\omega} - KL(q(\boldsymbol{\omega})\mid\mid p(\boldsymbol{\omega}))}_{\text{Same as ELBO in Section \ref{sec:BNN} }}\\
& \qquad + \underbrace{\int_{\boldsymbol{\omega}} q (\boldsymbol{\omega}) \log \mathcal{G}(\mathbf{H} \mid\mathbf{X},\boldsymbol{\omega})\mathit{d}\boldsymbol{\omega}}_{\text{New term, requires optimal prediction $\mathbf{H}$}} +\ \mathrm{const}.
\end{split}
\end{equation}
Next, using Monte Carlo integration and the dropout approximating distribution $q (\boldsymbol{\omega})$, this can be implemented as the standard objective loss of a dropout NN with an additional penalty term
\begin{equation}
\begin{array}{l}
\underbrace{-\sum_i \log p(\mathbf{y}_i \mid \mathbf{x}_i, \hat{\boldsymbol{\omega}}_i) + \mid\mid\boldsymbol{\omega}\mid\mid^2}_{\text{Equivalent to standard dropout loss}}\\
 \underbrace{ -\sum_i\left( \log \sum_{\mathbf{c}} u\left(\mathbf{h}_i, \mathbf{c}\right) p(\mathbf{y}_i=\mathbf{c}' \mid \mathbf{x}_i ,\hat{\boldsymbol{\omega}}_i)\right)}_{\text{Our additional utility-dependent penalty term}}
\end{array}
\end{equation}
where $\hat{\boldsymbol{\omega}}_i \sim q_{\boldsymbol{\theta}}(\boldsymbol{\omega})$. We alternate between one-step minimisation with respect to $\boldsymbol{\theta}$ and setting $\mathbf{h}_i$ using Equation \eqref{eq:opt_task_log}.

This derivation is influenced by \citet{lacoste2011approximate}, in which a related objective was derived for a simple tractable model, and optimised by applying variational EM to separately optimise the two parameters. In this section we extended the derivation to solve issues of non-tractability for large complex models, allowing the ideas of \citet{lacoste2011approximate} to be applied in real-world applications.

We display our technique for both learning the parameter weights and the optimal prediction in Algorithm \ref{al:LCBNN_opt} (Appendix \ref{ap:algo}), where we perform MC dropout by drawing Bernoulli distributed random variables $\boldsymbol{\epsilon}$ and apply the re-parameterisation $\boldsymbol{\omega} = \boldsymbol{\theta} \text{diag}(\boldsymbol{\epsilon})$, where $\boldsymbol{\theta}$ are the approximating distribution parameters (weight matrices' means)  \cite{gal2016dropout}.

\section{Illustrative Example}\label{sec:illust}

To give intuition into our loss-calibrated BNN, we introduce a scenario where we are required to diagnose a patient as either having severe diabetes, having moderate diabetes or being healthy. We borrow inspiration from \citet{leibig2017leveraging}, where they successfully used BNNs to diagnose diabetic retinopathy. For our illustrative example we synthesise a simple one-dimensional data set (this is to be able to visualise quantities clearly), whereby we have simulated three blood test results per patient assessing three ranges of blood sugar levels. Each blood test type indicates a patient belonging to one of the three possible classes. In particular, high values for tests $\{0,1,2\}$ correspond to `Healthy', `Moderate' and `Severe' respectively. We refer to Figure \ref{fig:ill_train}.

In this medical example, our goal is to avoid false negatives, whilst being less concerned about false positives. Table \ref{tab:med} demonstrates the mapping from the costs of incorrect misdiagnoses to a task-specific utility function. As an example, the highest cost and therefore lowest utility is assigned for a patient who is misdiagnosed as being healthy when their condition is severe.
\begin{table}[h]
\vskip -0.15in
\caption{A demonstration of converting the cost of making errors into a utility function for the illustrative example.}\label{tab:med}
\small
$$\begin{array}{c|cc|c}
\textbf{Cost of Incorrect} & \multirow{2}{*}{\textbf{Prediction}} & \multirow{2}{*}{\textbf{True}}& \textbf{Utility}\\
\textbf{Diagnosis}&&&\textbf{Function}\\
\hline
$\textsterling$ 0 & $Healthy$ & $Healthy$ & 2.0 \\
$\textsterling$ 0 & $Mild$ & $Mild$ & 2.0 \\
$\textsterling$ 0 & $Severe$ & $Severe$ & 2.0 \\
$\textsterling$ 30 & $Severe$ & $Mild$ & 1.4 \\
$\textsterling$ 35 & $Mild$ & $Severe$ & 1.3\\
$\textsterling$ 40 & $Mild$ & $Healthy$ & 1.2\\
$\textsterling$ 45 & $Severe$ & $Healthy$ & 1.1\\
$\textsterling$ 50 & $Healthy$ & $Mild$ & 1.0\\
$\textsterling$ 100 & $Healthy$ & $Severe$ & 0.0\\ 
\end{array}$$
\vskip -0.4in
\end{table}

Although in this example we have designed a rather arbitrary utility to capture our preference, the utility function values in such applications will often be assigned according to requirements set by health organisations. As an example, \citet{leibig2017leveraging} compare their (non-calibrated) results to sensitivity and specificity thresholds set by the NHS (UK National Health Service).

\subsection{Baseline Models}
We compare our model to two other techniques that are used in the literature. One approach is to ignore the structure of the utility function and use a standard BNN model to infer the network weights independently (identical to a standard dropout network with MC sampling at test time). Samples from this model are then used to integrate over the utility function at test time to make optimal predictions, as in Equation \eqref{eq:opt_task_log}. This technique would rely on perfectly approximating the posterior over the weights in order to maximise the utility. 

We also compare to the common approach in the field, which weighs different classes in the cross entropy differently\footnote{See the definition in Appendix \ref{ap:cross_entropy}.}, allowing us to put emphasis on certain classes in the network loss. To select weights for the weighted cross entropy model, we must be careful to select values that aid in maximising the expected utility. It is important to point out that selecting these weights adds further parameters to be tuned and it is not always clear which values to choose for maximising the expected utility. Finally, as with the previous baseline, we take samples from this dropout network as well and integrate over the utility function to choose optimal predictions.

\subsection{Data and Results}
\begin{figure}[h!]
    \centering
     \includegraphics[width=0.3\textwidth]{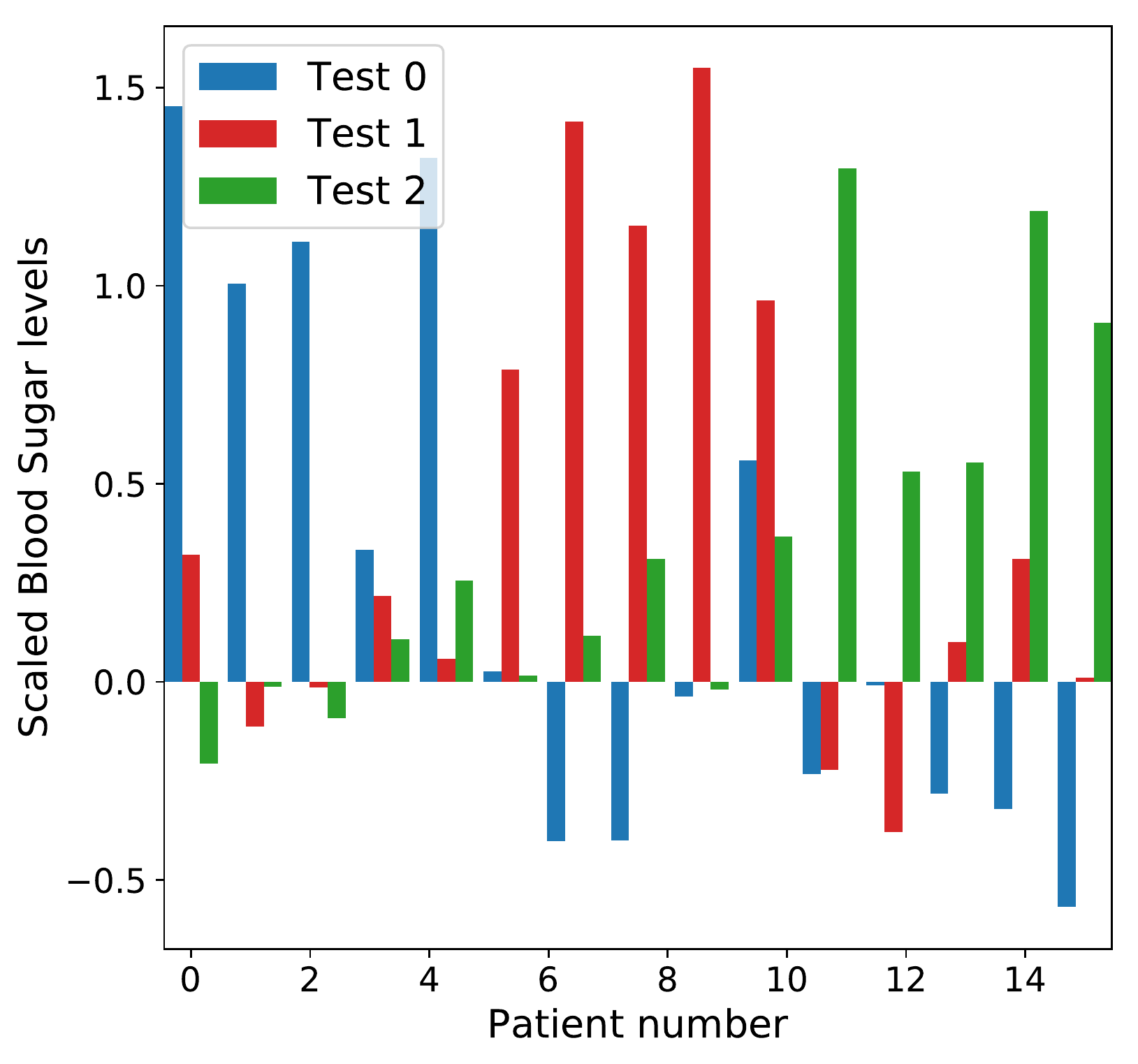}
    \caption{A sub-sample of scaled blood sugar levels for $15$ patients. Each patient diagnosis is based on these three-dimensional features.}\label{fig:ill_train}
\end{figure}

\begin{figure}[h!]
    \centering
        \includegraphics[width=\columnwidth]{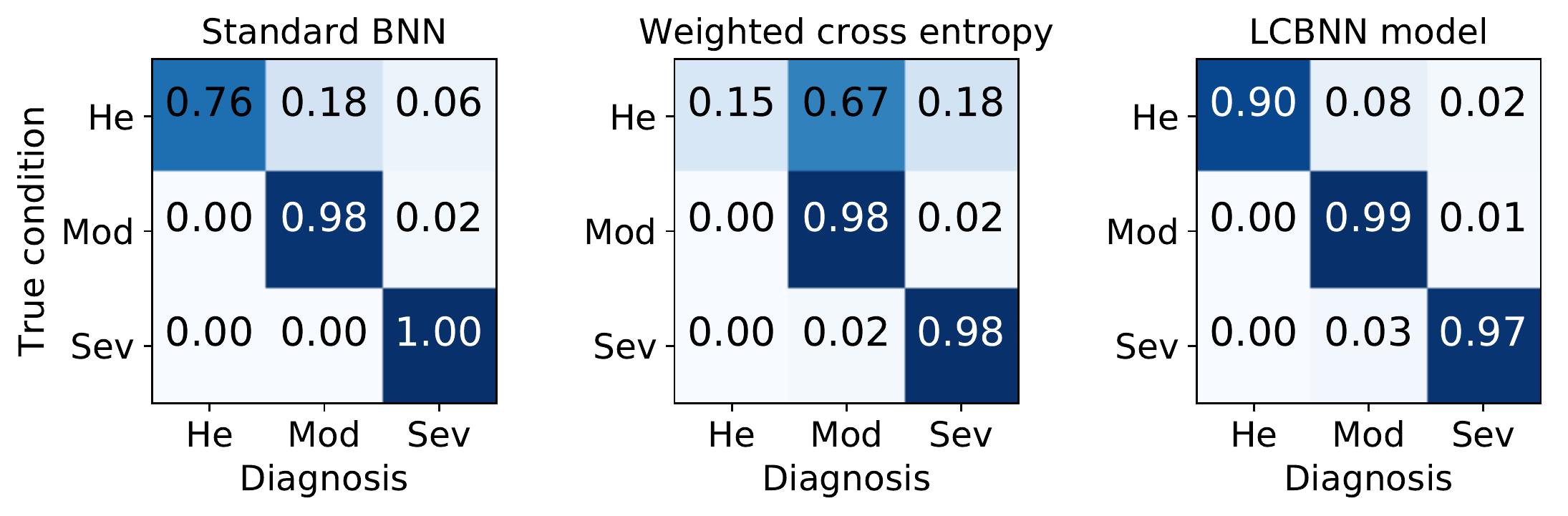}
    \caption{Left: Standard NN model. Middle: Weighted cross entropy model. Right: Loss-calibrated model. Each confusion matrix displays the resulting diagnosis when averaging the utility function with respect to the dropout samples of each network. We highlight that our utility function captures our preferences by avoiding false negatives of the `Healthy' class. In addition, there is a clear performance gain from the loss-calibrated model, despite the label noise in the training. This compares to both the standard and weighted cross entropy models, where there is a common failure mode of predicting a patient as being `Moderate' when they are `Healthy'.}\label{fig:ill_conf}
\end{figure}

We train the three models on the data shown in Figure \ref{fig:ill_train} and display our diagnosis predictions over a separate test set\footnote{Further experiment details are given in Appendix \ref{ap:med}.}.
The confusion matrices, in Figure \ref{fig:ill_conf}, demonstrate how the different models compare when making predictions.
For all the networks, we sample from the weights to get a distribution of network outputs. We then apply Equation \eqref{eq:opt_task_log} to make the diagnoses by averaging our outputs over the utility function. Due to the nature of the utility function, all networks avoid diagnosing unwell patients as healthy. Therefore we achieve the desired effect of avoiding false negatives. However, the key differences are in how each of the networks misclassify and how we enforce this behaviour.

The empirical gain calculated on the test data is higher for the standard NN than for that of the weighted cross entropy model. However, our loss-calibrated model outperforms both in achieving the highest empirical gain for this experiment when integrating over the utility.

\begin{figure}[t]
    \centering
    \begin{subfigure}[b]{0.4\textwidth}
        \includegraphics[width=\textwidth]{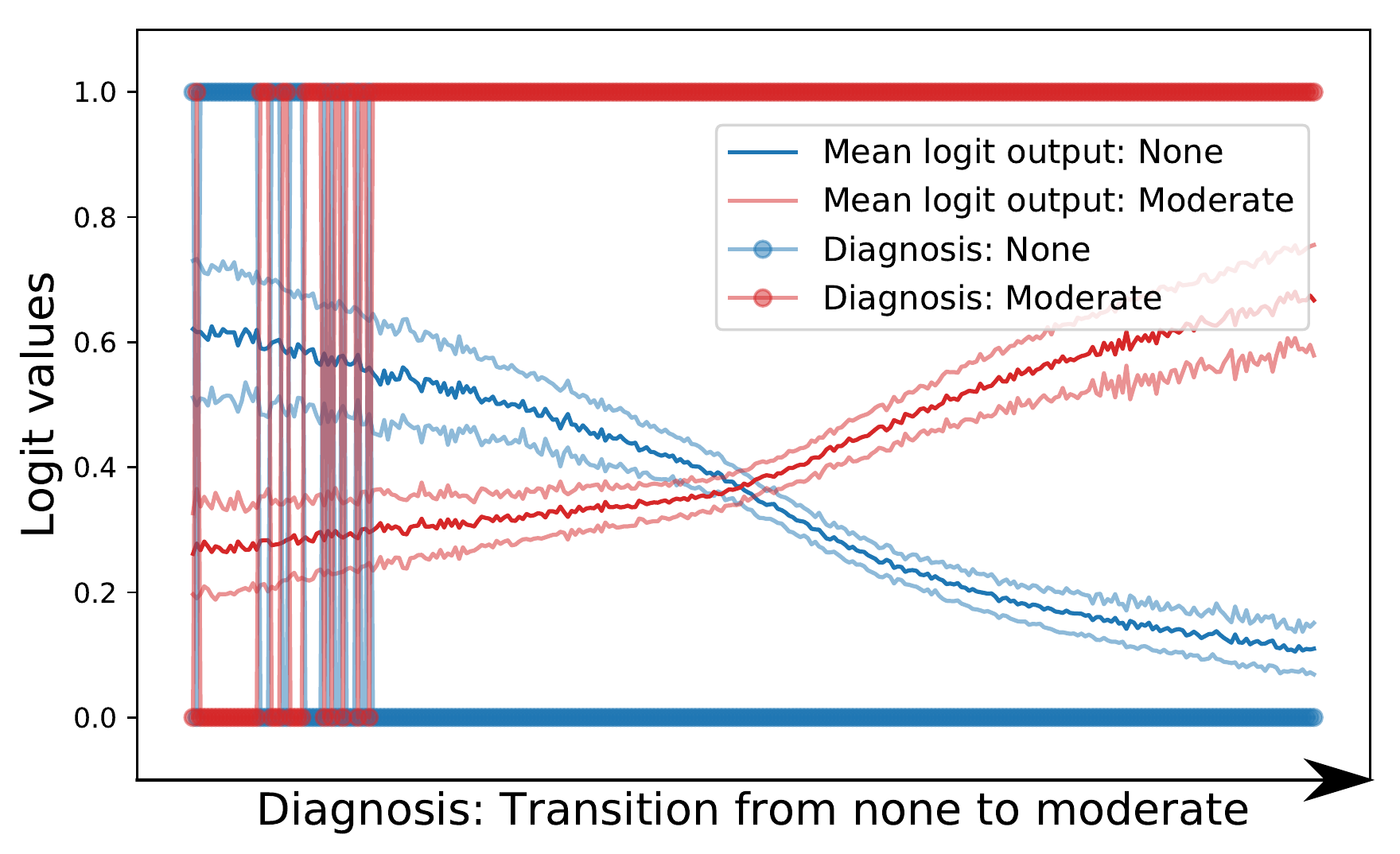}
        \caption{Weighted cross entropy}
        \label{fig:logits_we}
    \end{subfigure}
    ~ 
    \begin{subfigure}[b]{0.4\textwidth}
        \includegraphics[width=\textwidth]{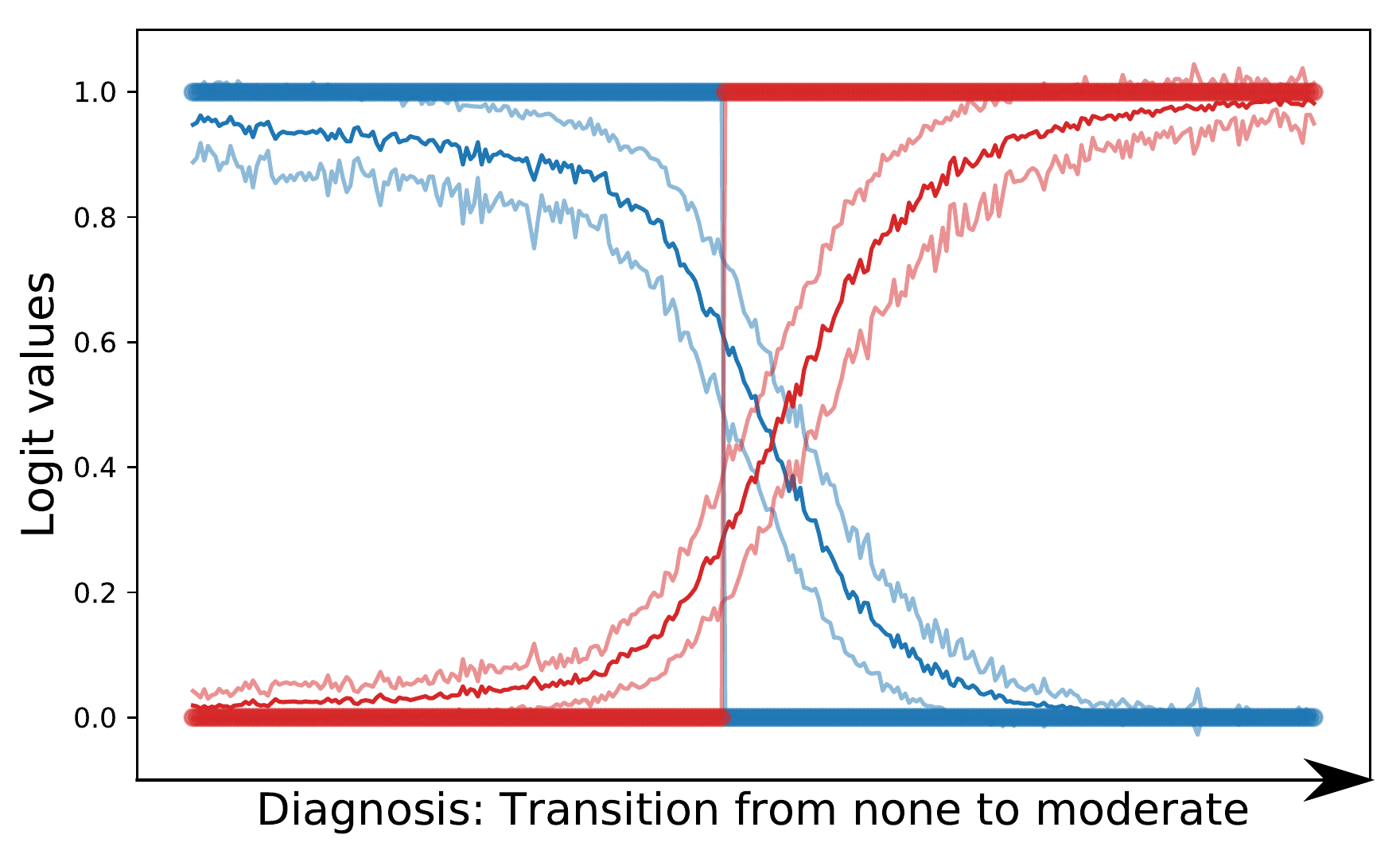}
        \caption{Loss-calibrated model}
        \label{fig:logits_lc}
    \end{subfigure}
    \vskip -0.1in
    \caption{We compare the behaviour of the loss-calibrated model (bottom) with the weighted cross entropy model (top) in order to demonstrate that weighting the cross entropy leads to over-fitting on erroneously labelled data. As we move from the feature space of a patient exhibiting no evidence of the disease to a patient with features indicating a moderate level of the disease, we display both the softmax outputs and the label predictions. The weighted cross entropy model favours diagnosing `Moderate' when integrating the utility over the model output. In contrast, the loss-calibrated model transitions smoothly. 
    }\label{fig:logits}
\end{figure}
\subsection{Uncertainty Behaviour}
In tandem with the results of Figure \ref{fig:ill_conf}, we offer an intuition in how the different models behave. Figure \ref{fig:logits} displays the network outputs from both the weighted cross entropy and the loss-calibrated model. We show the behaviour of each network during a transition from a patient with no diabetes to one that has moderate diabetes. We include the network softmax outputs, along with their standard deviations. Each figure also shows the optimal prediction, which is calculated by averaging the utility function over the softmax output samples.

Calibrating the network to take into account the utility leads to a smoother transition from diagnosing a patient as healthy to diagnosing them as having moderate diabetes. In comparison, weighting the cross entropy to avoid false negatives by making errors on the healthy class pushes it to `moderate' more often. This cautiousness, leads to an undesirable transition as shown in Figure \ref{fig:logits_we}. The weighted cross entropy model only diagnoses a patient as definitely being disease-free for extremely obvious test results, which is not a desirable characteristic. Much worse, we also see evidence of over-fitting to the training data occurring (not visible in the figure). The high weight on the moderate class penalises the model for not going exactly through the correspond moderate class $\mathbf{y}$ values, which leads to over-fitting. This over-fitting is not surprising given this model increases the weighting of erroneously-labelled noisy data (i.e.\ the noise model -- the likelihood -- has been changed).

\section{MNIST: Network Capacity and Label Corruption}\label{sec:mnist}

\begin{figure}[t]
    \centering
    \begin{subfigure}[b]{0.30\textwidth}
        \includegraphics[width=\textwidth]{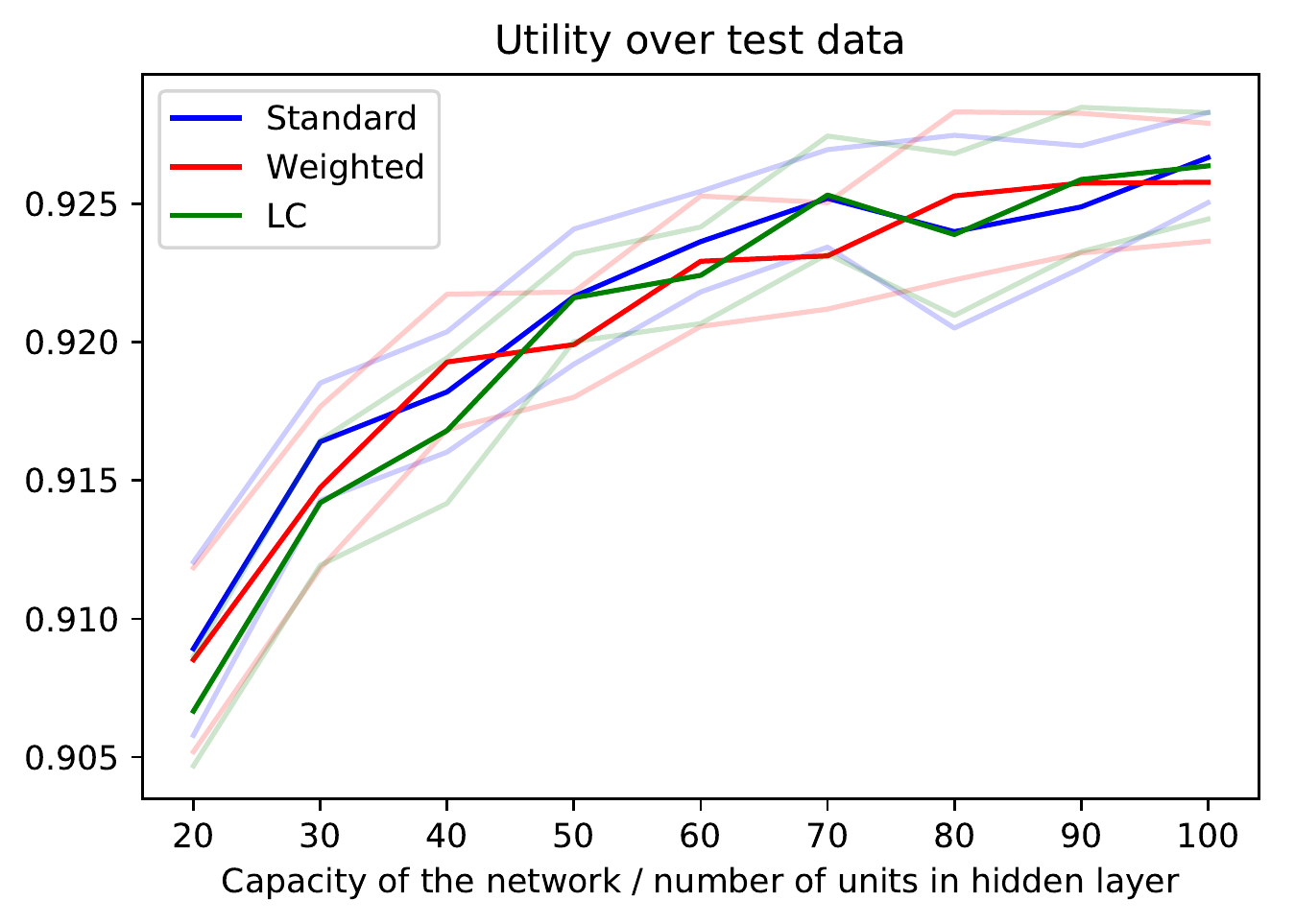}
        \caption{No label noise}
        \label{fig:no_label_noise}
    \end{subfigure}
    \qquad 
    \begin{subfigure}[b]{0.30\textwidth}
        \includegraphics[width=\textwidth]{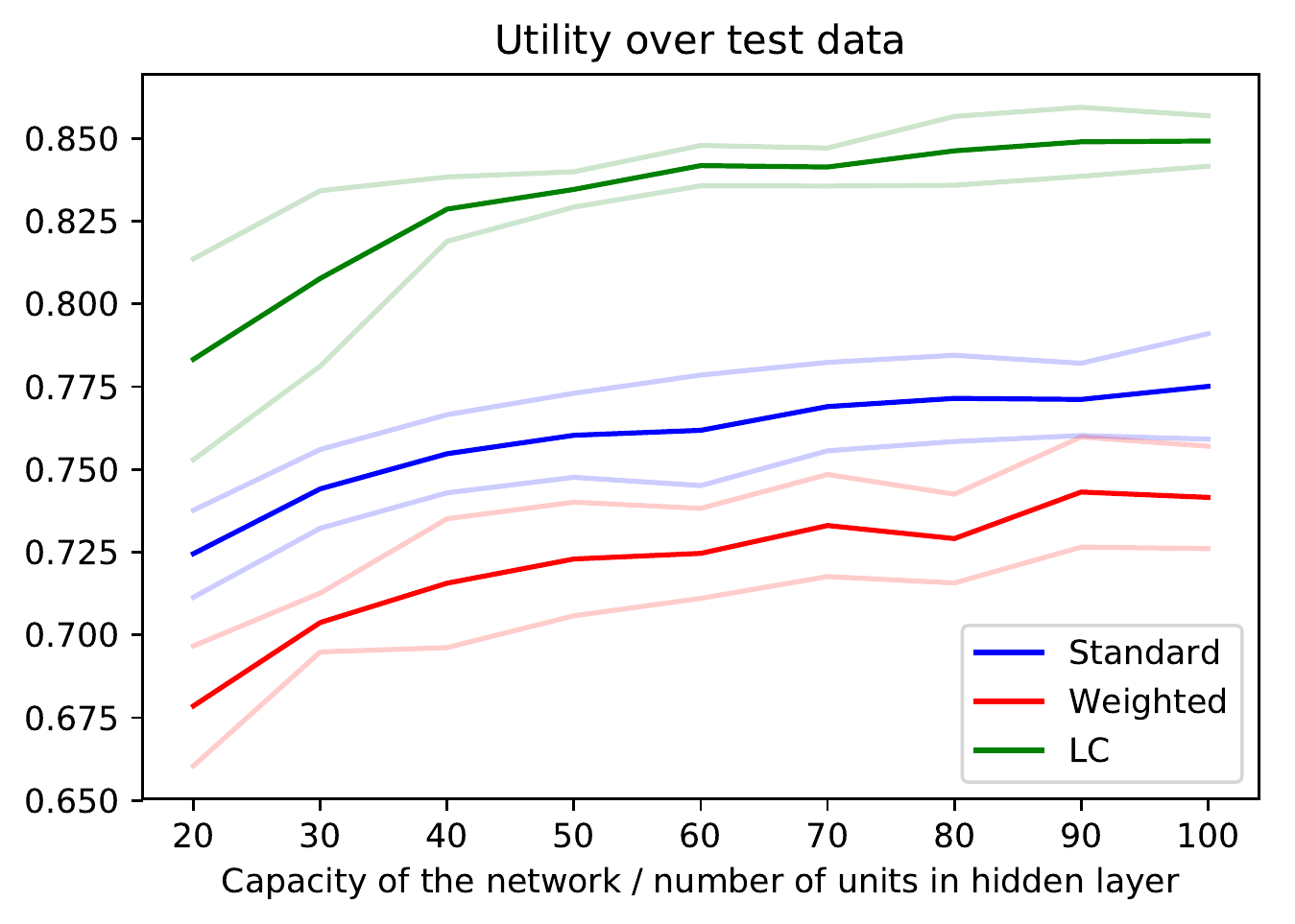}
        \caption{$50 \%$ uniformly corrupted label noise}
        \label{fig:05_label_noise}
    \end{subfigure}
    \caption{Each figure displays the expected utility over the test data as the size of the networks are increased. These results are calculated for $10$ random seeds and their corresponding one standard deviation bounds are included. For Figure \ref{fig:no_label_noise}, no label noise in training causes all the models to achieve similar utility over the uncorrupted test data. However, Figure \ref{fig:05_label_noise} shows that mislabelled training data can lead to severe over-fitting for our weighted cross entropy model, while our loss-calibrated model achieves the highest utility.}
 \label{fig:mnist_experiment}
\end{figure}

In many areas of machine learning, we come across scenarios where we are limited by the capacity of our model or by the quality of the data available. In this section, we demonstrate the use of our loss-calibrated model to target making optimal predictions for corrupted data with a limited capacity model. Furthermore, we show that the utility function forces the network to prioritise certain classes, in order to maximise the utility when the network is limited by the capacity. In addition, different noise levels are added to the labels to simulate a scenario where data contains corrupted observations. The corruption takes the form of a certain proportion of labels being reassigned according to a uniform distribution across all the classes. As an example, Figure \ref{fig:05_label_noise} displays an experiment, where a proportion of $50 \%$ of the labels are uniformly corrupted.

We apply our models to a modified version of the MNIST data set \cite{lecun1998gradient} and focus on maximising the utility for digits $\{3,8\}$, where our selection consists of classes that often contain ambiguities when trying to distinguish between them. For example, we highlight the similarities between digits $3$ and $8$. Our utility function is designed such that the maximum utility is achieved at $100 \%$ accuracy. However, the utility achieved in misclassifying digits $\{3,8\}$ is higher than a misclassification on the other digits. Therefore, the utility function 
encourages the network to focus on classes $\{3,8\}$ by  discouraging false negatives and accepting more false positives. Further details are given in Appendix \ref{ap:mnist_util}.

The clear result evident from Figure \ref{fig:mnist_experiment} is that weighting the cross entropy is a poor choice when facing an application that may have label noise. Figure \ref{fig:05_label_noise} shows that in addition to achieving a lower utility over the uncorrupted test data compared to our loss-calibrated model, the weighted cross entropy model also scores a worse utility than the vanilla model. Therefore, by increasing the weights applied to classes $\{3,8\}$, we actually suffer from a worse performance due to over-fitting, rather than encouraging the network to focus on these classes and achieve a higher utility.

Furthermore, Figure \ref{fig:no_label_noise} demonstrates that our utility-dependent lower bound does not have a detrimental effect on the performance, when the data set contains no label noise. This result is important as it indicates that our loss-calibrated network is the better choice of model for both scenarios.

\begin{figure*}[h!]
    \centering
    \begin{subfigure}[b]{1.2\columnwidth}
        \includegraphics[width=\columnwidth]{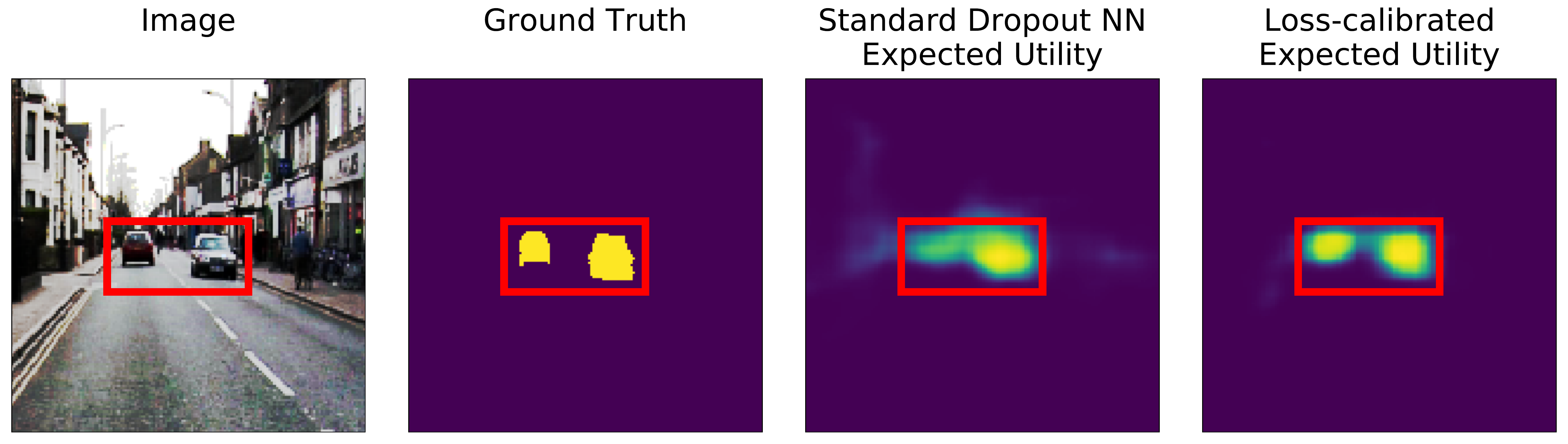}
        \caption{Segmentation of cars}
        \label{fig:car_1}
    \end{subfigure}
    \vskip 0.1in
    \begin{subfigure}[b]{1.2\columnwidth}
        \includegraphics[width=\textwidth]{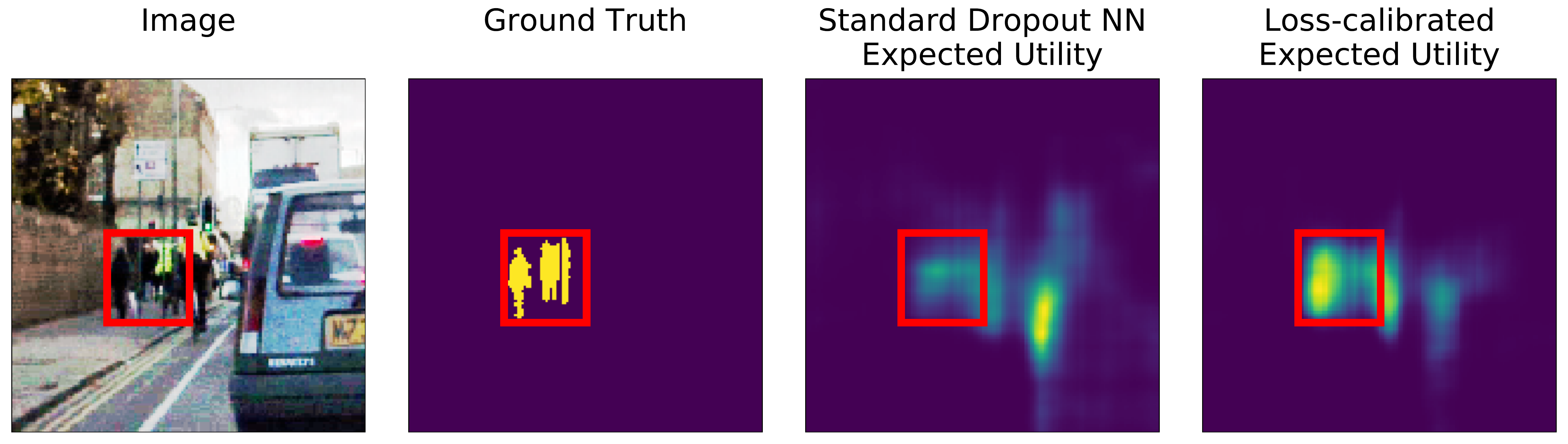}
        \caption{Segmentation of pedestrians}
        \label{fig:ped_1}
    \end{subfigure} 
        \caption{Utility maps comparing a standard Monte Carlo dropout NN with our loss-calibrated model using the SegNet-Basic architecture \cite{badrinarayanan2017segnet}. First column: a test image taken from the CamVid data set \cite{brostow2009semantic}. Second column: the corresponding ground truth for the car (\ref{fig:car_1}) and pedestrian (\ref{fig:ped_1}) classes. Third column: a utility map, given by the standard Bayesian SegNet Monte Carlo dropout implementation, showing the expected utility in assigning each pixel to the shown class. Yellow corresponds to a high gain, whereas blue correponds to a low gain. Fourth column:  a utility map given by our loss-calibrated model. In comparison to the standard model, our model produces a better behaved utility map, by placing a higher utility over the areas that contain pedestrians and cars.}\label{fig:seg_example_2}
\end{figure*}
\section{Per-Pixel Semantic Segmentation in Autonomous Driving}\label{sec:segnet}

In order to demonstrate that our loss-calibrated model scales to larger networks with real world applications, we display its performance on a computer vision task of per-pixel semantic segmentation using the data set CamVid \cite{brostow2009semantic}.
For this task, we design a utility function that captures our preferences for identifying pedestrians, cyclists and cars, over other classes such as trees, buildings and the sky (see Appendix \ref{ap:seg_util}). We then train our model and the baselines using the Bayesian SegNet-Basic architecture \cite{kendall2015bayesian}. Our backbone architecture is based on an implementation in Keras with a TensorFlow backend \cite{chollet2015keras,Konrad2016}, which consists of $9$ convolution layers and $5{,}467{,}500$ parameters.
The data set is split into $367$ training images, $101$ validation images and $233$ test images, all with $360 \times 480$ resolution.

Table \ref{tab:seg_res} displays results for our experiment. We highlight the importance of relying on the framework of Bayesian decision theory by displaying the results in two headings. The 
`Standard Prediction' gives the classification accuracy and the expected utility over the test data before any integration over the utility function. The `Optimal Prediction' shows the results of integrating over the utility. These results show that through this integration, we increase our expected utility over the test data across all models and better capture our preferences. Therefore this result advocates for the general use of combining BNNs with Bayesian decision theory.

\begin{savenotes}
\begin{table}[t]
\caption{Results over the test data for the Bayesian SegNet-Basic architecture. \textsc{Standard Pred.} corresponds to the classifications before integrating over the utility and \textsc{Optimal Pred.} corresponds to the results after the integration. We show that our utility on the test data greatly improves when assigning labels according to optimal prediction. Furthermore, our loss-calibrated model achieves the highest utility and highlights the benefits of our utility dependent lower bound.}
\vskip 0.1in
\label{tab:seg_res}
\begin{center}
\begin{small}
\begin{sc}
\begin{tabular}{ l  cc  cc }
 \toprule
 \parbox[t]{3mm}{\multirow{2}{*}{Models}}&\multicolumn{2}{c}{Standard Pred.}&\multicolumn{2}{c}{Optimal Pred.}\\ 
\cmidrule(lr){2-3}
\cmidrule(lr){4-5}
& Acc.& Exp. Util.&Acc. & Exp. Util.\\
\midrule
Standard&$81.1$	&	$0.619$  & $78.1$		& $0.676$\\
Weighted&$82.1$   &$0.633$   &$79.6$ &$0.682$\\
LC BNN&$82.5$     &$\mathbf{0.652}$	&$81.8$ &$\mathbf{0.685}$\\
\bottomrule
\end{tabular}
\end{sc}
\end{small}
\end{center}
\vskip -.1in
\end{table}
\end{savenotes}
\begin{table}[t]
\caption{Intersection of union (IOU) results to highlight how classes such as pedestrians and cars are prioritised over lower priority classes such as road, pavement and trees. The IOU results for the our loss-calibrated model clearly demonstrate how our model prioritises the higher utility classes, where these results are calculated from the optimal predictions.
}
\vskip 0.1in
\label{tab:IOU}
\setlength{\tabcolsep}{3.3pt}
\begin{center}
\begin{small}
\begin{sc}
\begin{tabular}{ l  c c c c c c }
 \toprule
 \parbox[t]{3mm}{\multirow{3}{*}{Models}}&\multicolumn{3}{c}{Low Utility}&\multicolumn{2}{c}{High Utility}& Mean\\
&\multicolumn{3}{c}{Classes IOU}&\multicolumn{2}{c}{Classes IOU}& IOU\\
\cmidrule(lr){2-4}
\cmidrule(lr){5-6}
\cmidrule(lr){7-7}
& Road & Pave. &Tree & Car & Ped.&All\\
\midrule
Standard&$0.85$	&	$0.65$  & $0.54$		& $0.28$ & $0.06$&$0.37$\\
Weighted&$0.86$   &$0.66$   &$0.55$ &$0.31$ &$0.09$&$0.40$\\
LC BNN&$0.86$     &$0.65$	&$0.54$ &$\mathbf{0.39}$ &$\mathbf{0.13}$&$0.42$\\
\bottomrule
\end{tabular}
\end{sc}
\end{small}
\end{center}
\vskip -.1in
\end{table}

In addition to highlighting the importance of the Bayesian-decision-theory-motivated evaluation scheme, Table \ref{tab:seg_res} shows our loss-calibrated model gives a performance boost over the current models. The benefits in incorporating the utility into the lower bound enables our model to achieve a higher utility than the models that are trained with no knowledge of the final application of the user.

Table \ref{tab:seg_res} also highlights the importance of evaluation metric choice. Accuracy gives equal weight to all classes, and cannot distinguish important classes from others. Sky pixels, which dominate the dataset, skew this metric unjustifiably. The expected utility metric, on the other hand, down-weighs sky pixels in the evaluation and up-weighs car and pedestrian pixel classifications.

Furthermore, Table \ref{tab:IOU} compares the intersection of union (IOU) for a subset of classes to highlight where the performance improvement lies. Our loss-calibrated model achieves similar IOUs for the classes with a lower utility, however our model demonstrates a relative increased performance on the more challenging higher priority classes shown in bold. We stress that the aim of this table is not to give state-of-the-art results but rather to demonstrate the performance of a calibrated model on the task of semantic segmentation, in comparison to standard approaches in the field.

Figures \ref{fig:car_1} and \ref{fig:ped_1} display utility maps over segments of test images (as in Figure \ref{fig:seg_example}). They give an intuition into how the labels for each model are assigned. 
These utility maps display the gain each model expects to receive, before knowledge of the ground truth is available. Our loss-calibrated model is able to capture sharper boundaries around the classes of interest. As an example, these sharper boundaries are especially obvious in Figure \ref{fig:car_1}, where the clear yellow circles, indicating high gain, are better defined for our loss-calibrated model than for the standard Bayesian SegNet.

\section{Conclusion}\label{sec:conc}
We have built a new utility-dependent lower bound for training BNNs, which allows our models to attain superior performance when learning an approximate distribution over weights for asymmetric utility functions. Furthermore, in relying on the framework of Bayesian decision theory, we have introduced a theoretically sound way of incorporating uncertainty and user preferences into our applications. The significance of our final segmentation experiment demonstrated the scalability of our loss-calibrated model to large networks with a real world application.

We highlighted the suitability of our approach for both medical applications and autonomous driving. Designing utility functions to encode assumptions corresponding to specific tasks not only provides better results over alternative methods, but also adds a layer of interpretability to constructing models. This clarity warrants further investigation into safety-critical applications.

\section*{Acknowledgements}
Adam D. Cobb is sponsored by the AIMS CDT (\url{http://aims.robots.ox.ac.uk}) and the EPSRC (\url{https://www.epsrc.ac.uk}). We thank NASA FDL (\url{http://www.frontierdevelopmentlab.org/#!/}) for making this collaboration possible and NVIDIA for granting us hardware. Furthermore, we also thank Richard Everett and Ivan Kiskin for extensive comments and feedback. 

\nocite{langley00}

\bibliography{bib_bib}
\bibliographystyle{icml2018}

\cleardoublepage
\section*{Appendix}

\appendix
\section{Bounding the Utility Function}\label{ap:bound}
Referring to \citet[Page~60]{berger1985statistical}, we bound the utility function to take positive values, such that $\log \mathcal{G}(\mathbf{H} \mid\mathbf{X},\boldsymbol{\omega})$ is defined for our loss-calibrated lower bound (Equation \ref{eq:expand_LB}).
Therefore throughout our experiments, we define a lower bound $M$, such that
$$M + \underset{\mathbf{h} \in \mathcal{H}}{\mathrm{inf}}\ \underset{\mathbf{y} \in \mathcal{Y}}{\mathrm{inf}}\ u\left( \mathbf{h}, \mathbf{y}\right) > 0.$$
We use the transformed utility function $u^*\left( \mathbf{h}, \mathbf{y}\right) = M + \underset{\mathbf{h} \in \mathcal{H}}{\mathrm{inf}}\ \underset{\mathbf{y} \in \mathcal{Y}}{\mathrm{inf}}\ u\left( \mathbf{h}, \mathbf{y}\right)$ for all experiments.

\section{Proof: KL Divergence Equivalence }\label{ap:proof}
To show that the maximisation of our loss-calibrated evidence lower bound is equivalent to minimising the KL divergence:\begin{equation*}\label{eq:KL}
KL(q\mid\mid \tilde{p}_h) = \log \left(\mathcal{G}(\mathbf{H} \mid\mathbf{X})\right) - \mathcal{L}(q(\boldsymbol{\omega}),\mathbf{H}),
\end{equation*}
where we have defined the probability distribution $\tilde{p}_h$ as
\begin{align}
\tilde{p}_h = \frac{p(\boldsymbol{\omega} \mid \mathbf{X,Y}) \mathcal{G}(\mathbf{H} \mid\mathbf{X},\boldsymbol{\omega})}{\mathcal{G}(\mathbf{H} \mid\mathbf{X})},\notag
\end{align}
we start with $KL(q\mid\mid \tilde{p}_h) = \int q \log \frac{q}{\tilde{p}_h}$:
\begin{align}
&KL(q\mid\mid \tilde{p}_h)\notag\\
&= \int_{\boldsymbol{\omega}} q(\boldsymbol{\omega})\log \left( \frac{q(\boldsymbol{\omega})}{\frac{p(\boldsymbol{\omega} \mid \mathbf{X,Y}) \mathcal{G}(\mathbf{H} \mid\mathbf{X},\boldsymbol{\omega})}{\mathcal{G}(\mathbf{H} \mid\mathbf{X})}}\right) \mathit{d}\boldsymbol{\omega}\notag \\
&= \int_{\boldsymbol{\omega}} q(\boldsymbol{\omega})\log \left( \frac{q(\boldsymbol{\omega})\mathcal{G}(\mathbf{H} \mid\mathbf{X})}{p(\boldsymbol{\omega} \mid \mathbf{X,Y}) \mathcal{G}(\mathbf{H} \mid\mathbf{X},\boldsymbol{\omega})}\right) \mathit{d}\boldsymbol{\omega}\notag
\end{align}
We can separate the above term into the log conditional-gain and the lower bound:
\begin{align}
&= \int_{\boldsymbol{\omega}} q(\boldsymbol{\omega})\log \left(\mathcal{G}(\mathbf{H} \mid\mathbf{X})\right)\mathit{d}\boldsymbol{\omega} \notag\\
&- \int_{\boldsymbol{\omega}} q(\boldsymbol{\omega}) \log \left(\frac{p(\boldsymbol{\omega} \mid \mathbf{X,Y}) \mathcal{G}(\mathbf{H} \mid\mathbf{X},\boldsymbol{\omega})}{q(\boldsymbol{\omega})}\right)\mathit{d}\boldsymbol{\omega}.\notag
\end{align}
As the conditional-gain $\mathcal{G}(\mathbf{H} \mid\mathbf{X})$ does not depend on $\boldsymbol{\omega}$ we recover:
\begin{align}
&KL(q\mid\mid \tilde{p}_h)\notag\\
& = \log \left(\mathcal{G}(\mathbf{H} \mid\mathbf{X})\right)\notag\\
&- \int_{\boldsymbol{\omega}} q(\boldsymbol{\omega}) \log \left(\frac{p(\boldsymbol{\omega} \mid \mathbf{X,Y}) \mathcal{G}(\mathbf{H} \mid\mathbf{X},\boldsymbol{\omega})}{q(\boldsymbol{\omega})}\right)\mathit{d}\boldsymbol{\omega}\notag \\
&= \log  \left(\mathcal{G}(\mathbf{H} \mid\mathbf{X})\right) - \mathcal{L}(q(\boldsymbol{\omega}),\mathbf{H})
\end{align}
as previously stated.

\section{Algorithm: LCBNN}\label{ap:algo}
Our implementation is shown in Algorithm \ref{al:LCBNN_opt}. We follow the same notation that was introduced in the paper.
\begin{algorithm}[H]\label{al:LCBNN_opt}
\caption{LCBNN optimisation}\label{al:LCBNN_opt}
\begin{algorithmic}[1]\label{al:LCBNN_opt}
  \scriptsize
  \STATE Given dataset $\mathcal{D} = \{\mathbf{X},\mathbf{Y}\}$, utility function $u(\mathbf{h},\mathbf{y})$ and set of all possible labels $\mathcal{C}$
\STATE Define learning rate schedule $\eta$  
  \STATE Randomly initialise weights $\boldsymbol{\omega}$ 
  \REPEAT
  \STATE Sample $S$ index set of training examples
  \FOR{$i \in S$}
  	\FOR{$t$ from $1$ to $T$}
  	\STATE Sample Bernouilli distributed random variables $\boldsymbol{\epsilon}^t \sim p(\boldsymbol{\epsilon})$ \COMMENT{for each each $\mathbf{x}_i$ we sample T dropout masks $\boldsymbol{\epsilon}^t$}
  	\STATE $\mathbf{y}_i^t =  \mathbf{f}^{g(\boldsymbol{\omega},\boldsymbol{\epsilon}^t)}(\mathbf{x}_i)$ \COMMENT{Perform a stochastic forward pass with the sampled dropout mask $\boldsymbol{\epsilon}^t$ and $\mathbf{x}_i$}
  	\ENDFOR 
  \STATE $\mathbf{h}_i^* \leftarrow \underset{\mathbf{h} \in \mathcal{H}}{\mathrm{argmax}}\ \frac{1}{T}\sum_t u(\mathbf{h},\mathbf{y}_i^t)$ \COMMENT{Choose class $\mathbf{h} = \mathbf{c}\in \mathcal{C}$ which maximises average utility}
 \ENDFOR
  \STATE Calculate the derivative w.r.t. $\boldsymbol{\omega}$:
  \begin{equation*}
  \begin{split}
  \nabla\boldsymbol{\omega} \leftarrow & -\frac{1}{T}\sum_{i \in S} \frac{\partial}{\partial \boldsymbol{\omega}}\log p\left(\mathbf{y}_i \mid \mathbf{f}^{g(\boldsymbol{\omega},\boldsymbol{\epsilon}_i)}(\mathbf{x}_i)\right)\\
  & \qquad + \frac{\partial}{\partial \boldsymbol{\omega}} KL\left(q(\boldsymbol{\omega})\mid\mid p(\boldsymbol{\omega})\right)\\
  & \qquad - \frac{1}{T}\sum_{i \in S}\frac{\partial}{\partial \boldsymbol{\omega}}\log \mathcal{G}\left(\mathbf{h}^*_i \mid \mathbf{x}_i,\boldsymbol{\omega}\right)
\end{split}
  \end{equation*}
with $\boldsymbol{\epsilon}_i \sim p(\boldsymbol{\epsilon})$ a newly sampled dropout mask for each $i$.
 \STATE Update $\boldsymbol{\omega}$: 
 $\boldsymbol{\omega} \leftarrow \boldsymbol{\omega} + \eta \nabla\boldsymbol{\omega}$

 \UNTIL{$\boldsymbol{\omega}$ has converged}
\end{algorithmic}\label{al:LCBNN_opt}
\end{algorithm}
\section{Baseline Model: Weighted Cross Entropy}\label{ap:cross_entropy}
In order to overcome large class imbalances in data, when training a NN, the common technique is to apply a weighting $\alpha_i$ to each class in the cross entropy loss as follows:
$$\text{loss} = \sum_{i=1}^{C} \alpha_i\  p(\mathbf{y} = c_i \mid \boldsymbol{\omega}, \mathbf{x}),$$
where for each class $i$, we have a corresponding weight $\alpha_i$ to indicate the size of its contribution to the cross entropy loss. The term, $p(\mathbf{y} = c_i \mid \boldsymbol{\omega}, \mathbf{x})$, is the categorical-softmax defined in Equation \ref{eq:softmax}.

\section{Experiment: Illustrative Example}\label{ap:med}
\subsection{Data}
To provide further details about the data from our illustrative example, we refer back to Figure \ref{fig:ill_train}. We display a sub-sample of patients, where for each patient the three test results are displayed in a bar chart.
These test results correspond to the input data that a doctor might use to make their diagnosis. 

\begin{figure}[h!]
\centering
        \includegraphics[width=0.2\textwidth]{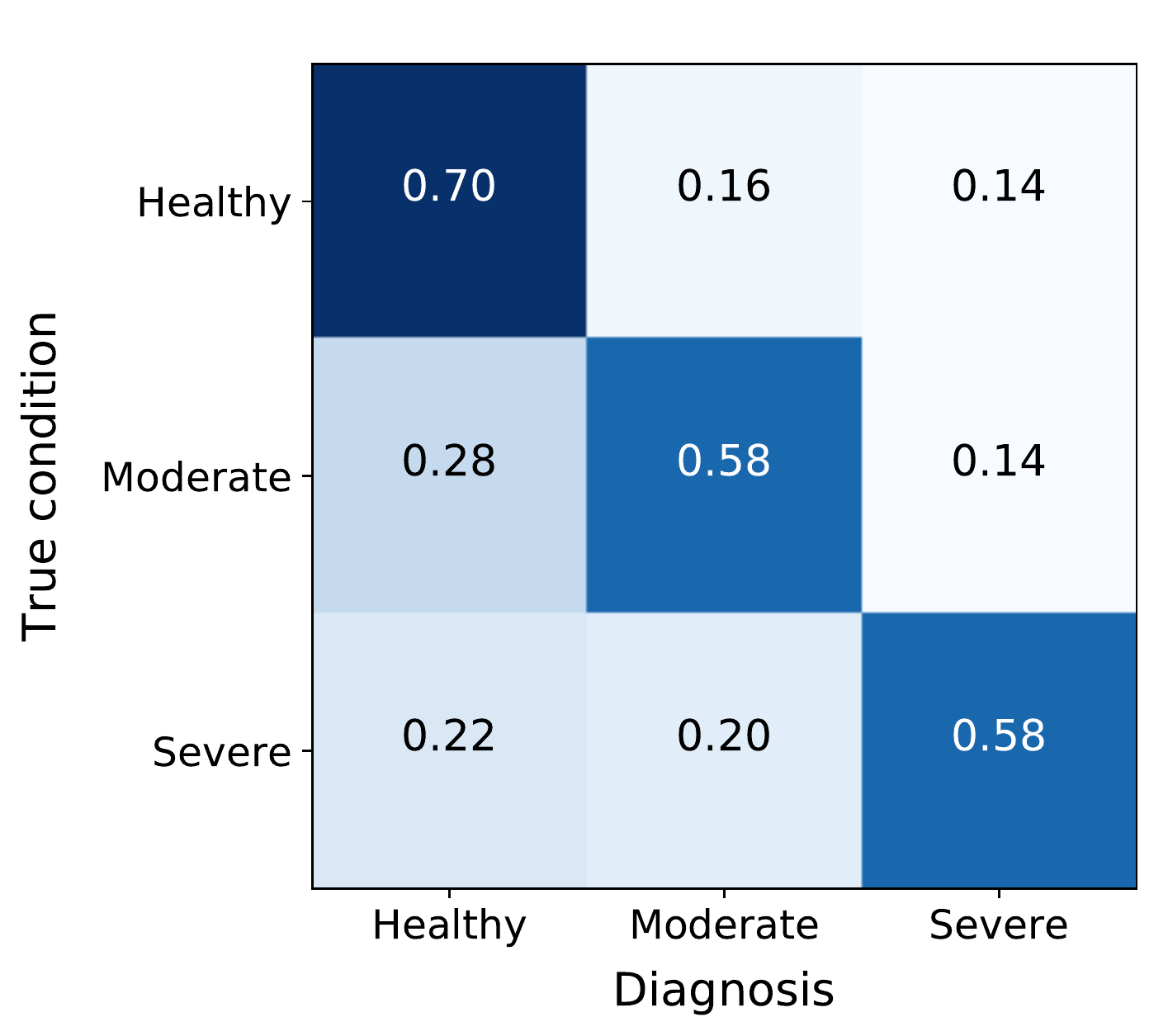}
        \caption{Corrupted observations}
        \label{fig:ill_label}
    \end{figure}

Figure \ref{fig:ill_label} shows how the training set consists of mislabelled data and simulates patients being misdiagnosed. 
The values in each block of the matrix correspond to the proportion of the total $150$ patients. 
\subsection{Architecture}
For all three models, we use the same architecture consisting of one hidden layer with $20$ units. We apply the same regularisation to all layers and apply a dropout of $0.2$.

\subsection{Baseline}
In order to complement the utility function, we assign weights to match the relative values in assigning labels to each class. Therefore, through empirical experimentation, we select weights $[1.0,2.0,2.0]$ to correspond to `Healthy', `Moderate' and `Severe' respectively.

\section{Experiment: MNIST}\label{ap:mnist_util}
\subsection{Utility Function}
The utility function prescribes $0.3$ for false positives of the digits $\{3,8\}$ and $0.0$ for false positives for all other digits. Correct classifications are given a maximum utility of $1.0$.

\subsection{Architecture}
All models consist of one hidden layer with ReLU activations. We set dropout to $0.2$ and the lengthscale to $0.01$. 

\subsection{Additional Results}

All experiments are trained on $2{,}500$ images and tested on $10{,}000$ uncorrupted images. 

\begin{figure}[h]
    \centering
    \begin{subfigure}[b]{0.24\textwidth}
        \includegraphics[width=\textwidth]{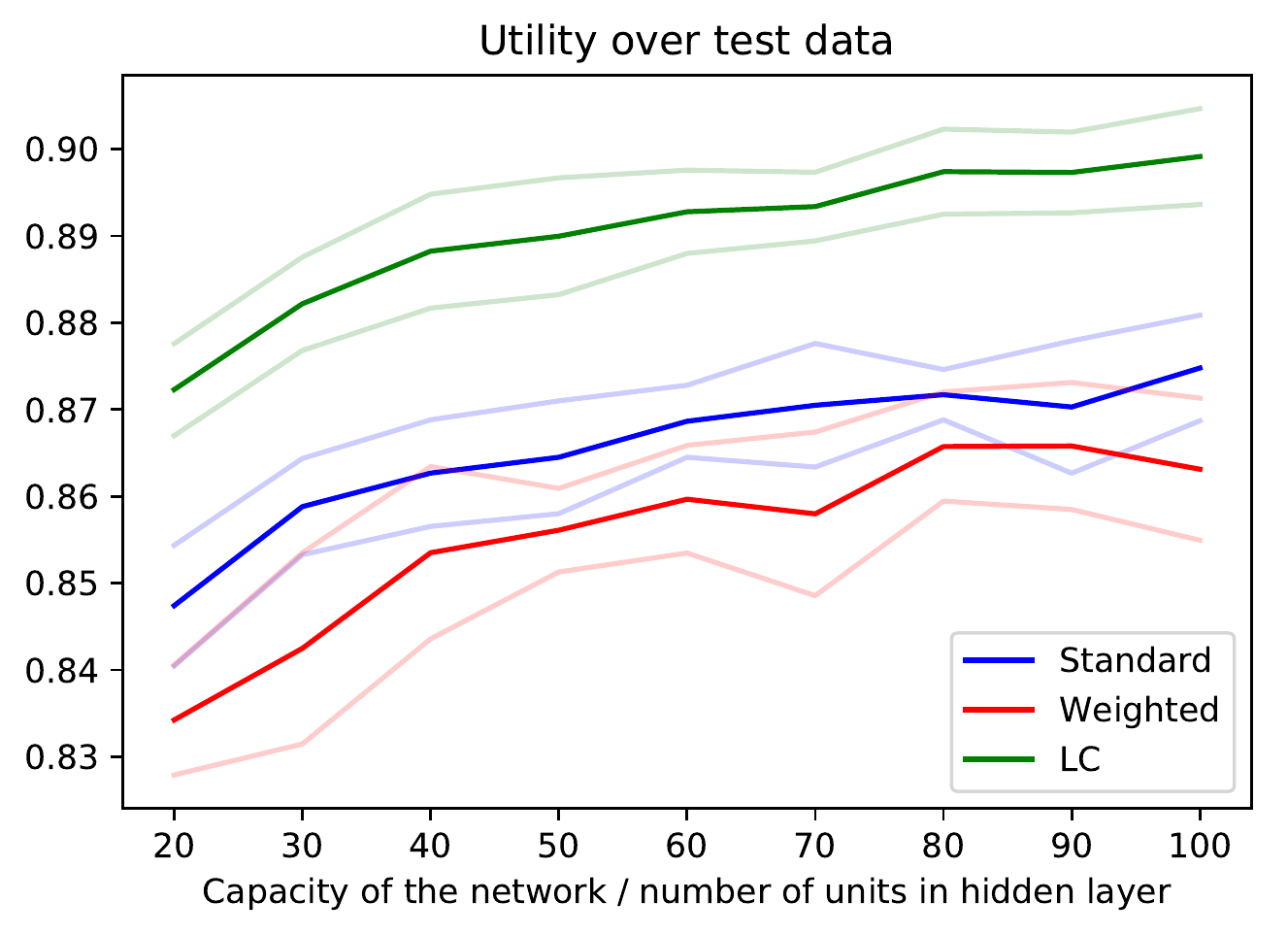}
        \caption{$25 \%$ uniformly corrupted label noise}
        \label{fig:025_label_noise}
    \end{subfigure}
    \begin{subfigure}[b]{0.24\textwidth}
        \includegraphics[width=\textwidth]{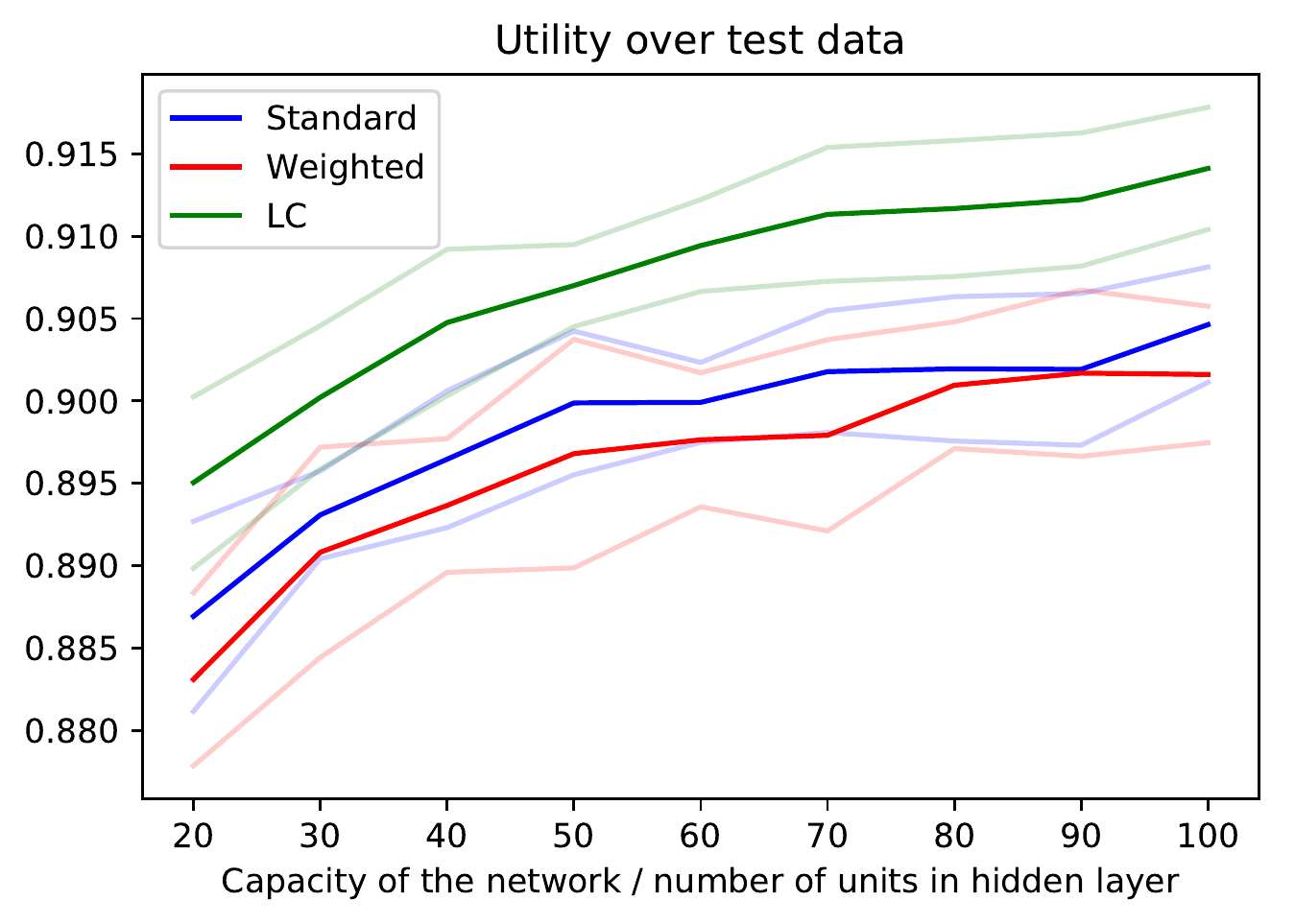}
        \caption{$10 \%$ uniformly corrupted label noise}
        \label{fig:01_label_noise}
    \end{subfigure}
    \caption{These results are calculated for $10$ random seeds, where they were trained on $2500$ data points and tested on $10000$. The models converge on utility as we reduce the noise in the training data. In the main paper, Figure \ref{fig:mnist_experiment} shows the two extremes of when there is no label noise and where there is $50 \%$ uniformly corrupted label noise.} \label{fig:mnist_extra}
\end{figure}

We include additional results in Figure \ref{fig:mnist_extra} from two more experiments with different noise levels of $25 \%$ and $10 \%$ on the training data.
\section{Experiment: Segmentation}\label{ap:seg_util}

\subsection{Utility}
\begin{table}[h!]
\caption{Our utility function for the per-pixel semantic segmentation task.} 
\label{tab:seg_util}
\begin{tiny}
\[\arraycolsep=1.5pt\def\arraystretch{1.5}
\begin{array}{lc|cccccccccccc}
&&\multicolumn{12}{c}{\textbf{True}}\\
&\textbf{Utility} &\textbf{S.} &\textbf{B.} &\textbf{Po.} &\textbf{R.} &\textbf{Pa.}&\textbf{T.} &\textbf{S.}&\textbf{F.} &\textbf{C.} &\textbf{Pe.}&\textbf{C.}&\textbf{U.}\\
\hline
\parbox[t]{2mm}{\multirow{12}{*}{\rotatebox[origin=c]{90}{\textbf{Prediction}}}}
&\textbf{Sky}        	& \mathbf{0.8} & 0.0 & 0.0 & 0.0 & 0.0 & 0.0 & 0.0 & 0.0 & 0.0 & 0.0 & 0.0 & 0.0 \\
&\textbf{Building}  	& 0.0 & \mathbf{0.8} & 0.0 & 0.0 & 0.0 & 0.0 & 0.0 & 0.0 & 0.0 & 0.0 & 0.0 & 0.0 \\
&\textbf{Pole}        	& 0.2 & 0.2 & \mathbf{0.8} & 0.2 & 0.2 & 0.2 & 0.2 & 0.2 & 0.2 & 0.2 & 0.2 &0.2\\
&\textbf{Road}       	& 0.2 & 0.2 & 0.2 & \mathbf{0.8} & 0.2 & 0.2 & 0.2 &  0.2 & 0.2 & 0.2 & 0.2 &0.2\\
&\textbf{Pavement}	&  0.2 & 0.2 & 0.2 & 0.2 & \mathbf{0.8} & 0.2 & 0.2 & 0.2 & 0.2 & 0.2 & 0.2 & 0.2\\
&\textbf{Tree} 			& 0.2 & 0.2 & 0.2 & 0.2 & 0.2 & \mathbf{0.8} & 0.2 & 0.2 & 0.2 & 0.2 &0.2 &0.2\\
&\textbf{Sign} 			& 0.2 & 0.2 & 0.2 & 0.2 & 0.2 & 0.2 & \mathbf{0.8} & 0.2 & 0.2 & 0.2 & 0.2 & 0.2\\
&\textbf{Fence} 		& 0.2 & 0.2 & 0.2 & 0.2 & 0.2 & 0.2 & 0.2 & \mathbf{0.8} &0.2 & 0.2 & 0.2 & 0.2\\
&\textbf{Car} 			& 0.4 & 0.4 & 0.4 & 0.4 & 0.4 & 0.4 & 0.4 & 0.4 & \mathbf{0.8} &0.4 &0.4 &0.4\\
&\textbf{Pedestrian}& 0.4 & 0.4 & 0.4 & 0.4 & 0.4 & 0.4 & 0.4 & 0.4 & 0.4 & \mathbf{0.8} & \mathbf{0.8} &0.4\\
&\textbf{Cyclist} 		& 0.4 & 0.4 & 0.4 & 0.4 & 0.4 & 0.4 & 0.4 & 0.4 & 0.4 & \mathbf{0.8} & \mathbf{0.8} &0.4\\
&\textbf{Unlabelled}&  0.2 & 0.2 & 0.2 & 0.2 & 0.2 &  0.2 & 0.2 & 0.2 & 0.2 & 0.2 & 0.2 &\mathbf{0.8}\\
\end{array}
\]
\end{tiny}
\end{table}
We refer to Table \ref{tab:seg_util} for our defined utility function.
To encourage fewer false negatives for cars, pedestrians and cyclists, we assign a higher utility ($0.4$) for false positives relative to the other categories. Maximum utility ($0.8$) is achieved for correct labelling and lowest utility ($0.0$) is given to errors in predicting the sky and buildings to discourage the network from focusing on these classes.

\subsection{Architecture}
We follow the \emph{Bayesian SegNet-Basic Central Four Encoder-Decoder} architecture \citep{kendall2015bayesian}, where a dropout of $0.5$ is inserted after the central four encoder and decoder units.

\end{document}